\pdfoutput=1
\documentclass[runningheads]{llncs}
\usepackage{graphicx}
\usepackage{CJKutf8}
\usepackage{indentfirst}
\usepackage[utf8]{inputenc}
\input{zhwinfonts}
\begin{document}


\title{Applications of BERT Based Sequence Tagging Models on Chinese Medical Text Attributes Extraction}
\author{Gang Zhao\quad
Teng Zhang\quad 
Chenxiao Wang\quad 
Ping Lv\quad 
Ji Wu\quad
}
\authorrunning{ }
\institute{Tsinghua-iFlytek Joint Laboratory\\
\email{\{zhaogang\_ee, wangcx18,luping\_ts,wuji\_ee\}@mail.tsinghua.edu.cn}\\
\email{zhangteng1887@gmail.com}\\}
\maketitle
\begin{abstract}

    We convert the Chinese medical text attributes extraction task into a sequence tagging or machine reading comprehension task. Based on BERT pre-trained models, we have not only tried the widely used LSTM-CRF sequence tagging model, but also other sequence models, such as CNN, UCNN, WaveNet, SelfAttention, etc, which reaches similar performance as LSTM+CRF. This sheds a light on the traditional sequence tagging models. Since the aspect of emphasis for different sequence tagging models varies substantially, ensembling these models adds diversity to the final system. By doing so, our system achieves good performance on the task of Chinese medical text attributes extraction (subtask 2 of CCKS 2019 task 1).

\end{abstract}

\begin{CJK*}{UTF8}{gbsn}
\CJKindent  
\title{融合的bert序列标注模型\\在医疗文本属性抽取任务上的应用}
%
%
\author{赵 刚\quad
张 腾\quad 
王晨骁\quad 
吕 萍\quad 
吴 及\quad
}
\authorrunning{ }
%
\institute{清华大学科大讯飞联合实验室\\
\email{\{zhaogang\_ee, wangcx18,luping\_ts,wuji\_ee\}@mail.tsinghua.edu.cn}\\
\email{zhangteng1887@gmail.com}\\}
\maketitle              
\begin{abstract}
我们将医疗文本属性抽取任务转换为序列标注任务和阅读理解任务，在bert预训练模型的基础上，除了业界认可的LSTM+CRF序列标注模型之外，我们尝试了CNN，UCNN，WaveNet，Self\_attention等多种序列标注模型，达到了和LSTM+CRF一样的性能效果，对传统的序列标注任务有一定的启发意义。不同的的序列标注模型侧重点会有较大差异，增加了系统的多样性，通过模型融合，我们在医疗文本属性抽取任务上取得了不错的性能。

\end{abstract}

%
%
%
\section{任务定义及数据集}
结合数据源“癌症医疗影像检查与结论”的内容及特点，定义若干与癌症医疗病历相关的目标字段，如癌症原发部位，病灶大小和癌症转移部位等。原发部位是某种癌症最先发生的组织或者器官，如肺癌原发于左肺上叶；病灶大小是原发部位的大小，通常以最大直径或者大小直径表示；转移部位是癌症从最先发生的组织或器官转移到的其他组织或器官。

训练及测试数据分为四部分：1)900条非目标场景的标注数据，2)100条目标场景的标注数据，3)1000条各个场景的非标注数据，4）400条目标场景的标注数据作为最终评测的测试集。

\section{系统概述}
\begin{figure}
\begin{center}
\includegraphics[width=0.5\textwidth]{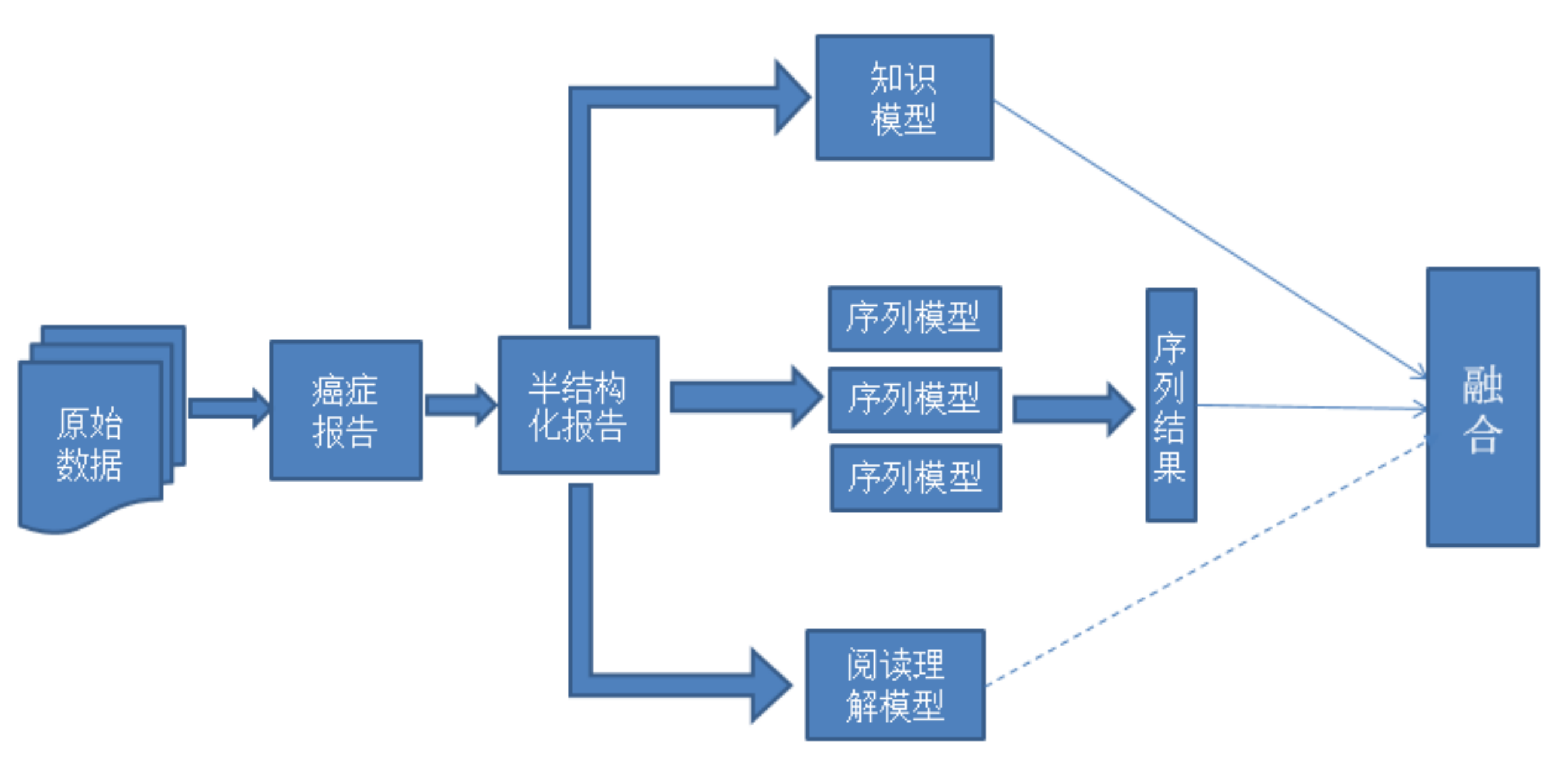}
\caption{信息流图} \label{fig2}
\end{center}
\end{figure}

信息流图如图\ref{fig2}所示，首先过滤掉一些非癌症报告，例如心脏超声，良性肿瘤等，依据描述内容，将报告分成“印象”和“所见”两部分，详见2.1文本预处理；然后使用融合的bert序列标注模型预测属性值，详见第4章序列标注模型；我们也做了一些阅读理解模型相关实验，内容及结果会在第5章阅读理解模型中有所介绍，但此次评测中没有加入阅读理解模型。

\subsection{文本预处理}

影像报告中会有一些心脏彩超，良性肿瘤的报告，不属于我们要处理内容。我们会使用规则过滤掉这一部分内容，如果文本中没有提到“癌症”、“肿瘤”或者“转移”等关键词，不进入下一阶段。

\begin{figure}
\begin{center}
\includegraphics[width=0.4\textwidth]{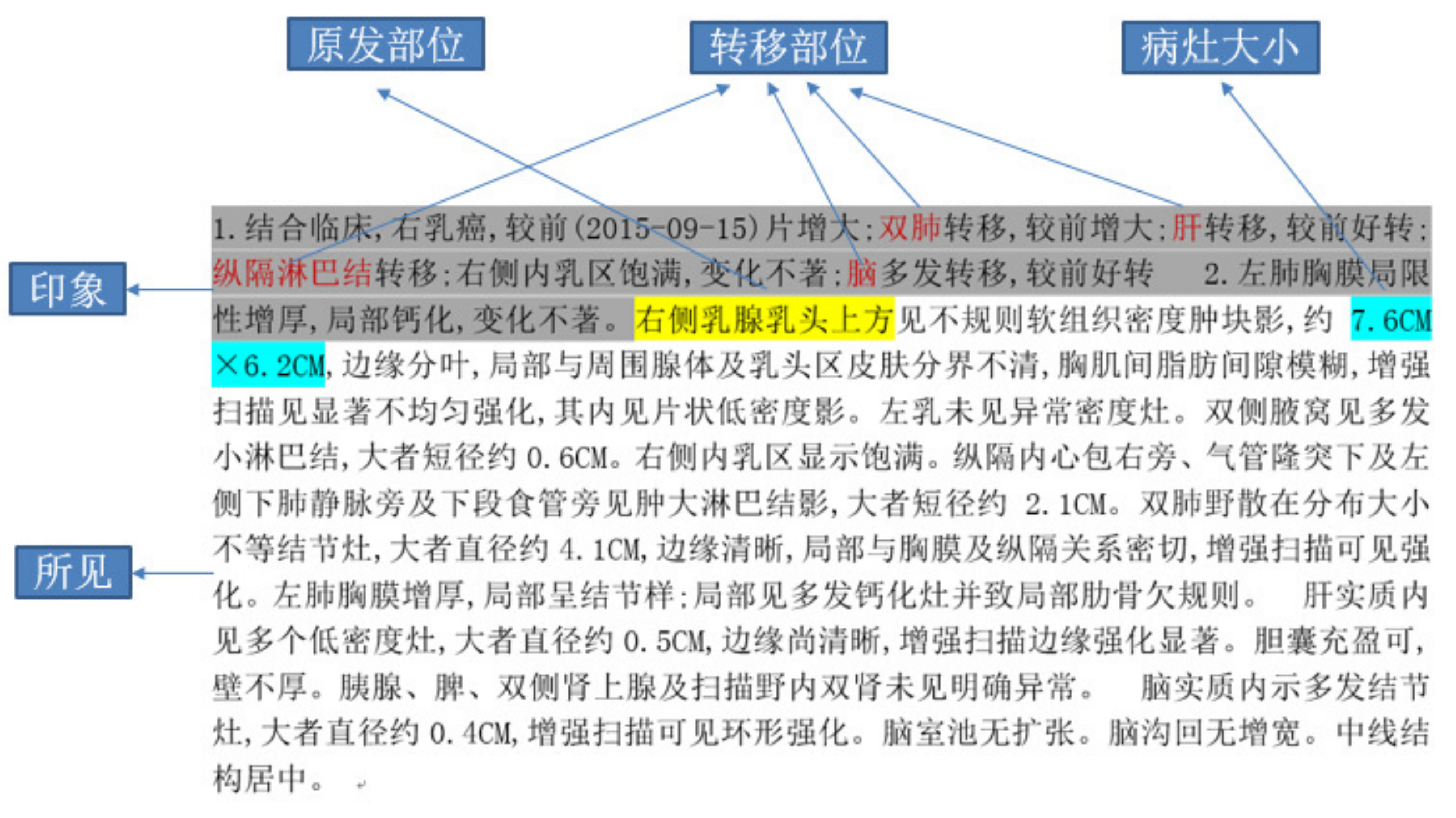}
\caption{文本预处理} \label{fig12}
\end{center}
\end{figure}

我们发现这些医疗报告有某种固定结构，如图\ref{fig12}所示，1、灰色背景是结论性的文本，我们称之为“印象”，转移部位和原发部位一般会在这里；2、白色背景是描述性的文本，原发部位和病灶大小一般会在这里（一般在第一句话里），我们称之为“所见”。首先使用规则将印象和所见的第一句话提取出来，作为一个半结构化报告进入后续系统。

\section{bert预训练}

预训练语言模型在很多任务中证明是有效的，包括句子级别任务（语言推理），字符级别任务（命名实体识别和问答系统）等都取得了业界最好的性能指标。预训练语言模型使用两种策略应对下游任务，一种是与下游任务相关，下游任务的特征会加入到预训练中来，例如ELMo\cite{ref_proc14}；一种是与下游任务无关的，只使用自然语言本身的特征（句子级别和字符级别特征），不需要定制下游任务的特征，例如bert\cite{ref_proc3}。自从bert问世以来，已经在十一项自然语言任务中取得了最好成绩，后来出现了很多以bert为基础的模型持续刷新各项自然语言处理任务的榜单。本文主要阐述了bert预训练模型在医疗文本上的应用（序列标注模型和阅读理解模型）。

使用bert模型主要包括两个步骤pre\_train和fine\_tune，pre\_train用来学习文本的单词特征、句法特征和语义特征，fine\_tune用来学习下游任务标签与文本表示相关特征。

\begin{figure}
\begin{center}
\includegraphics[width=0.4\textwidth]{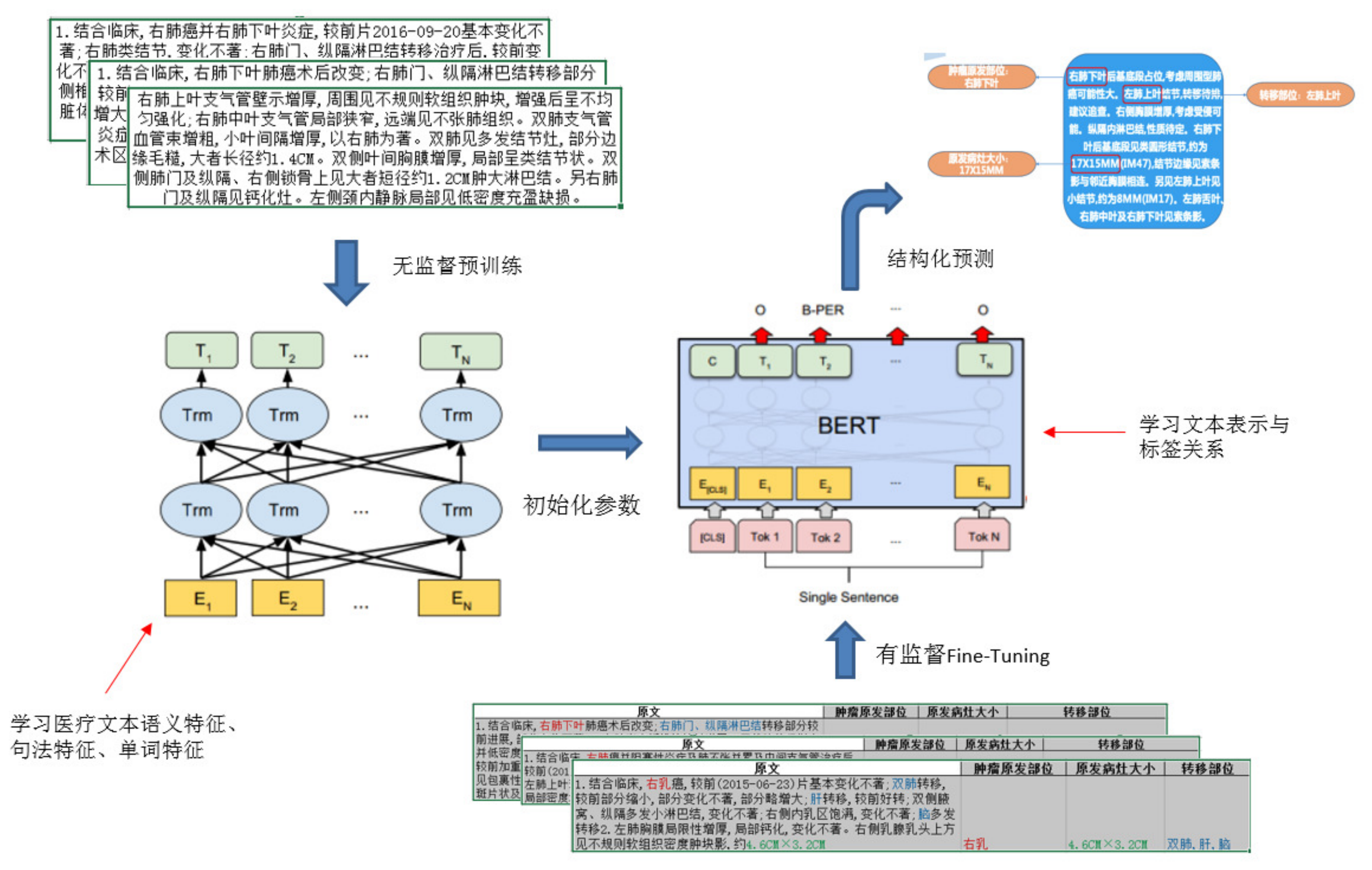}
\caption{bert预训练及Fine\_tune} \label{fig3}
\end{center}
\end{figure}

\section{序列标注模型}

\subsection{条件随机场}
\begin{figure}
\begin{center}
\includegraphics[width=0.4\textwidth]{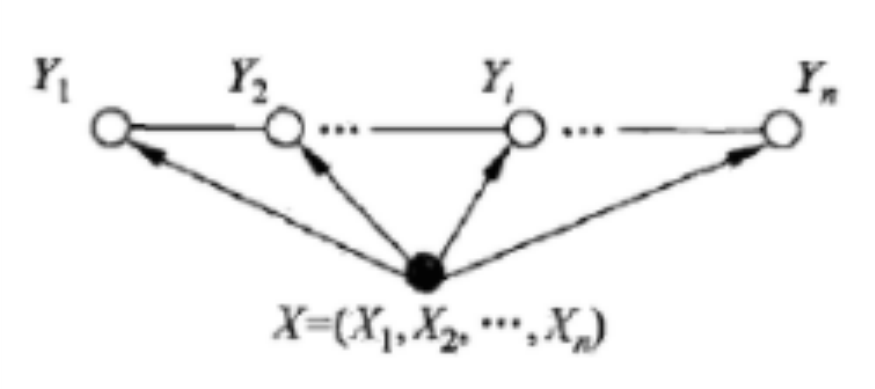}
\caption{条件随机场} \label{fig8}
\end{center}
\end{figure}

 条件随机场(Conditional Random Fields，CRF)由Lafferty等人\cite{ref_proc15}于2001年提出，结合了最大熵模型和隐马尔可夫模型的特点，是一种无向图模型（如图\ref{fig8}所示），近年来在分词、词性标注和命名实体识别等序列标注任务中取得了很好的效果。条件随机场是一个典型的判别式模型，其联合概率可以写成若干势函数联乘的形式，其中最常用的是线性链条件随机场。若让x=(x1，x2，…xn) \quad 表示被观察的输入数据序列，y=(y1，y2，…yn)表示一个状态序列。文本概率分布可分解为单词条件概率的乘积和。

\subsection{长短期记忆模型}
\begin{figure}
\begin{center}
\includegraphics[width=0.4\textwidth]{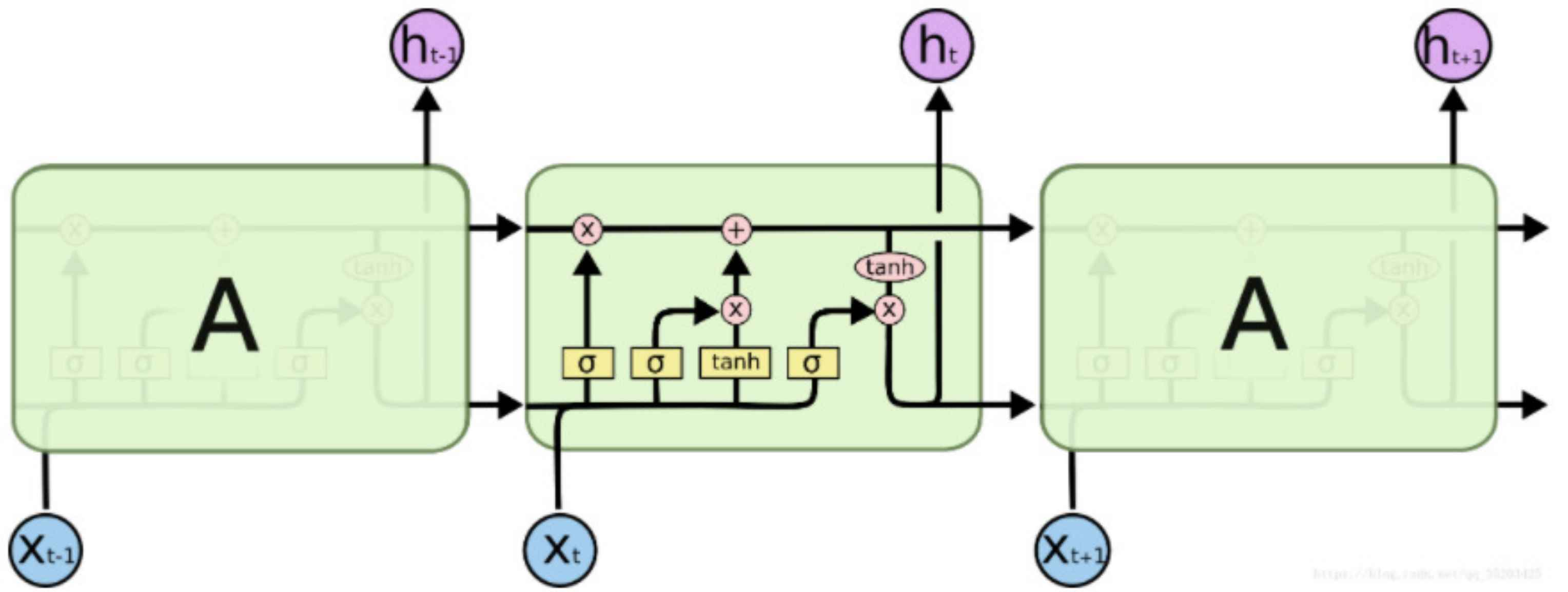}
\caption{长短时记忆} \label{fig7}
\end{center}
\end{figure}

长短期记忆（Long short\-term memory，LSTM）是一种特殊的RNN，主要是为了解决长序列训练过程中的梯度消失和梯度爆炸问题。简单来说，就是相比普通的RNN，LSTM能够在更长的序列中有更好的表现，结构如图\ref{fig7}所示。这里我们使用双向LSTM提取特征，后面接一个线性模型降维，最后接一个softmax层进行解码。

\subsection{卷积神经网络}

\begin{figure}
\begin{center}
\includegraphics[width=0.5\textwidth]{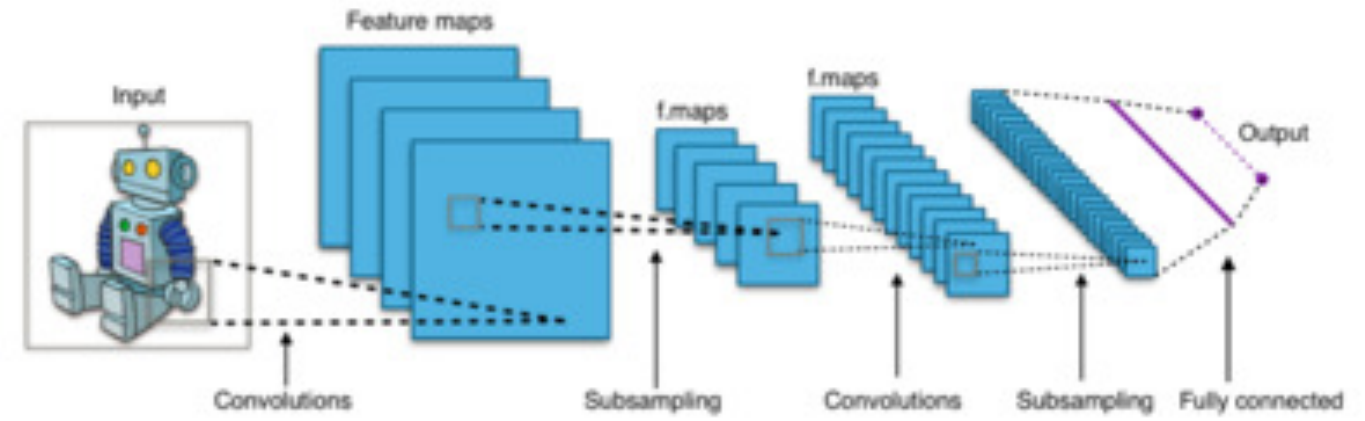}
\caption{卷积神经网络} \label{fig9}
\end{center}
\end{figure}

卷积神经网络（Convolutional Neural Network, CNN）是一种前馈神经网络，它的人工神经元可以响应一部分覆盖范围内的周围单元，由一个或多个卷积层和顶端的全连通层组成，一般也包括池化层（pooling layer）。与其他深度学习结构相比，卷积神经网络在图像和语音识别方面能够给出更好的结果，近年卷积神经网络在自然语言处理领域也得到越来越多的应用。也可以使用反向传播算法进行训练，相比较其他深度前馈神经网络，卷积神经网络需要考量的参数更少，结构如图\ref{fig9}所示。
\cite{ref_proc4}证明了深层的卷积神经网络能很好的提取特征。

我们这里设置卷积核大小分别为[2,3,4]，卷积核数量分别为[128,128,128]，卷积神经网络之后加一个全连接层，再过一个CRF层进行解码。

\subsection{self\_attention}

Attention机制最早在图像中使用，后来在机器翻译中被证明有效。self\_attention \quad 能够重点关注到有用的输入信息，作为特征提取器被广泛使用。attention函数可以看成是将query和一些key-value对映射成一个输出，输出是value的加权和，权重由query和key的某种相似性来度量。

\begin{figure}
\begin{center}
\includegraphics[width=0.5\textwidth]{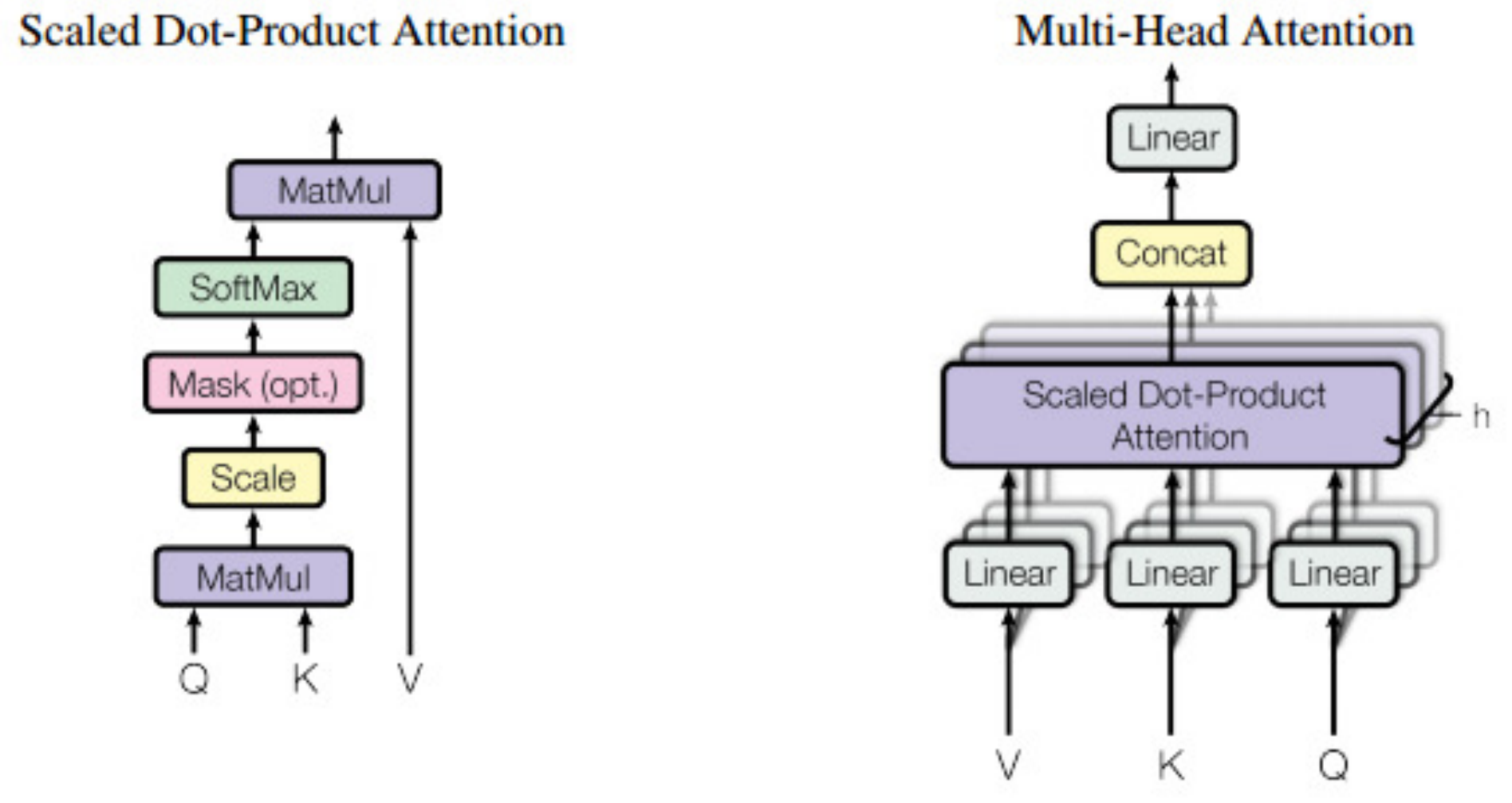}
\caption{Multi\_head attention} \label{fig5}
\end{center}
\end{figure}

我们使用了Multi-head attention\cite{ref_proc2}作为特征抽取器，结构如图\ref{fig5}所示，公式如下所示：
\begin{equation}
Attention(Q,K,V)=softmax(\frac{QK^T}{\sqrt d_k})V
\end{equation}
\begin{equation}
MultiHead(Q,K,V) = Concat(head_1,\dots,head_h)W^O
\end{equation}
\begin{equation}
head_i = Attention(QW_i^Q,KW_i^K,VW_i^V)
\end{equation}

这里我们设置线性层维度为256，Multi-head维度为16。

\subsection{WaveNet}

\begin{figure}
\begin{center}
\includegraphics[width=0.5\textwidth]{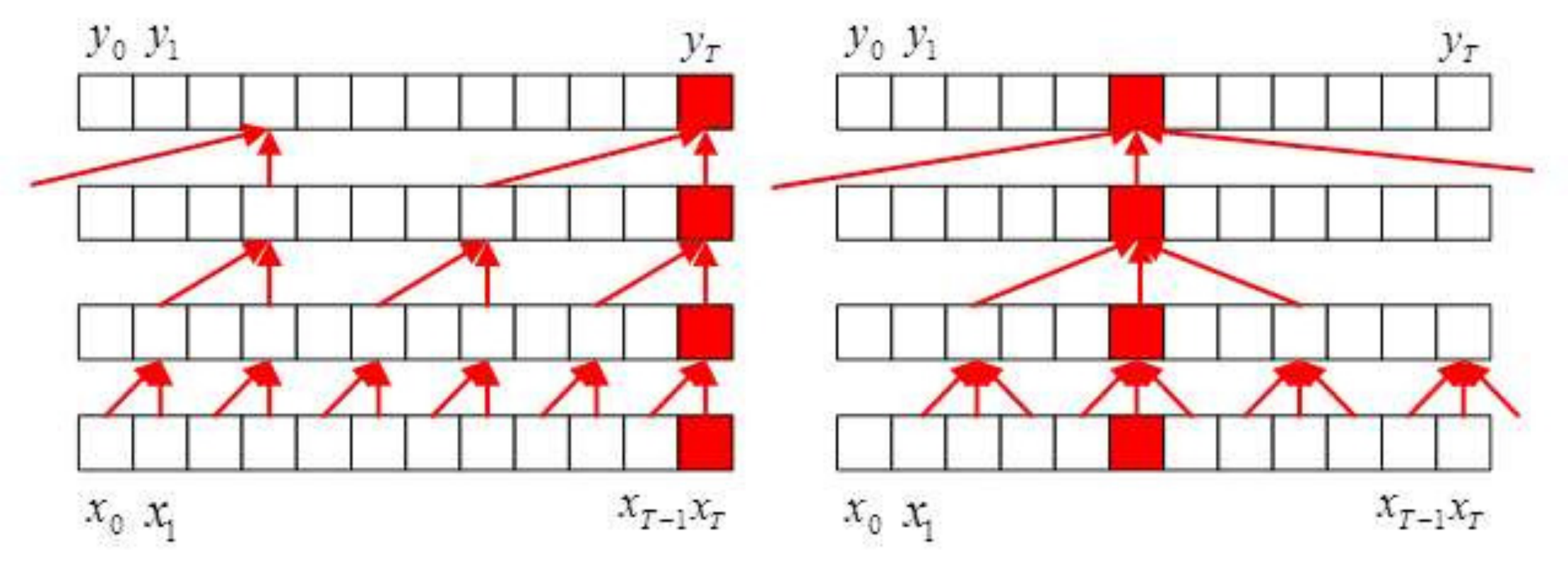}
\caption{左：扩张卷积结构，右：双向扩张卷积结构} \label{wavenet}
\end{center}
\end{figure}

近年来，无监督表示学习方法在自然语言处理领域取得了巨大的成功\cite{ref_proc7,ref_proc8,ref_proc9}。这些方法的基本策略是首先使用大规模文本语料库训练神经网络，然后在下游的小规模任务上对模型参数进行微调。自回归模型是常用的无监督表示学习方法之一。自回归语言模型试图用自回归的方法估计文本语料库的概率分布。具体来说，给定文本序列X=(x1,…,xT)，自回归语言模型将文本概率分布分解为单词条件概率的前向乘积 或者后向乘积 。WaveNet是谷歌deepmind在2016年发表的用于语音合成的自回归模型[4]。相比于文本序列来说，语音数据具有更长的序列长度，相当于至少每秒16000个样本。因此，WaveNet虽然脱胎于传统的卷积神经网络，但由于其特殊的扩张卷积结构，它可以建模更长范围内的文本相关性。
扩张卷积结构如图\ref{wavenet}左所示。假设卷积核大小为2，三层的扩张卷积结构可以建模长度为8的序列信息。作为对比，如果使用传统的卷积神经网络，卷积核大小为2时，需要7层卷积结构才能长度为8的序列信息。

使用WaveNet进行有监督的文本序列建模时，存在如下两个问题：
\begin{enumerate}
    \item WaveNet作为典型的自回归模型，无法完成文本序列的双向建模；
    \item WaveNet是一种无监督表示学习方法，无法直接应用于有监督任务。
\end{enumerate}

为了解决这两个问题，我们对原始WaveNet结构进行了如下修改。首先，我们将WaveNet中的扩张卷积结构修改为如图\ref{wavenet}右所示。双向扩张卷积结构融合了传统卷积神经网络的双向建模能力和扩张卷积结构的长时建模能力。然后，我们将网络输出修改为有监督任务中的标注序列，从而把WaveNet从无监督的序列生成任务移植到有监督的序列分类任务中来。 

\subsection{UCNN}
\begin{figure}
\begin{center}
\includegraphics[width=0.5\textwidth]{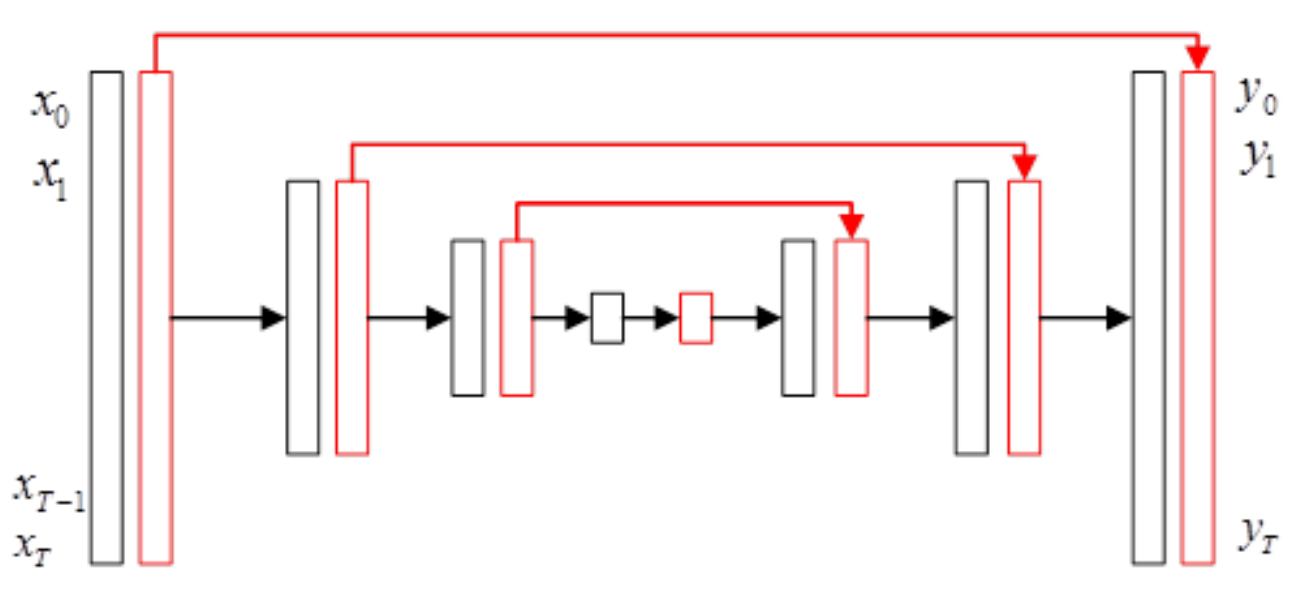}
\caption{UCNN结构} \label{ucnn}
\end{center}
\end{figure}

在图像分析领域，充分利用来自不同空间尺度的信息是图像处理的有效方式，这样做的主要原因是因为图像中物体的大小会随着与观察者的距离而改变。对于文本序列这样的时间序列来说，时间上的缩放属性并不十分明显，但是我们认为不同的时间尺度在文本序列分析中仍然非常重要，比如文本按照不同的时间尺度，可以切分成字、词、短语、句子、段落等等。在卷积神经网络结构中，残差网络\cite{ref_proc11}首次建立起相邻卷积层之间的直接联系，将底层的特征信息直接传递到顶层，是对图像多尺度空间信息充分利用的典型案例。在图像语义分割任务中广泛使用的U-net\cite{ref_proc12}更加直观的对图像的多尺度空间信息进行了融合。我们将传统U-net的网络结构进行了相应改造，以满足文本序列处理的需求，如图\ref{ucnn}所示。首先，我们将传统U-net中的二维卷积操作、二维下采样操作和二维上采样操作分别替换为一维卷积操作、一维下采样操作和一维上采样操作。然后，我们将网络输出修改为有监督任务中的标注序列，从而使UCNN网络能够使用于文本序列分类任务。
\subsection{Attention-LSTM}
\begin{figure}
\begin{center}
\includegraphics[width=0.5\textwidth]{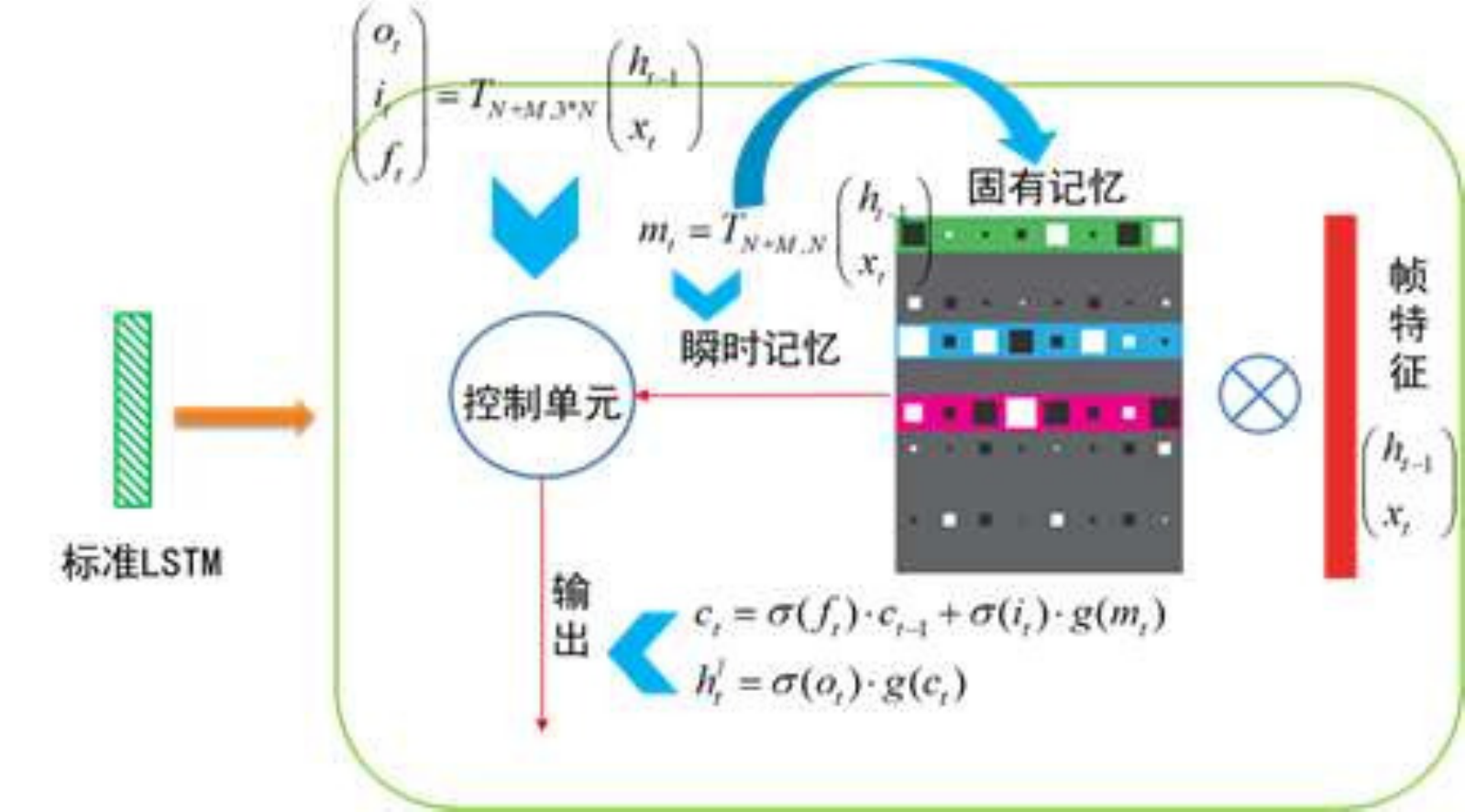}
\caption{标准长短时记忆模型中的记忆机制} \label{attention_lstm1}
\end{center}
\end{figure}

\begin{figure}
\begin{center}
\includegraphics[width=0.5\textwidth]{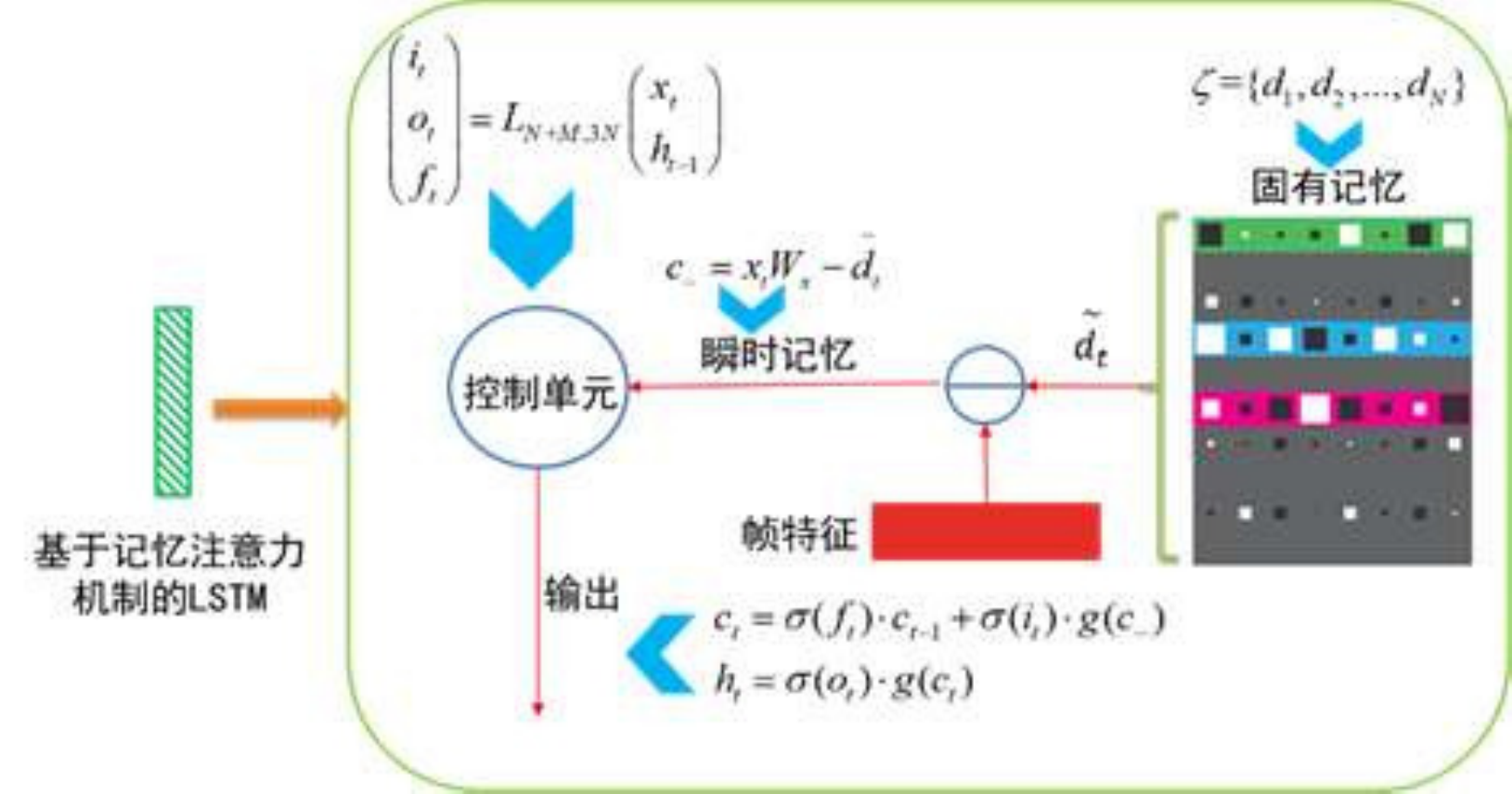}
\caption{基于记忆注意力机制的长短时记忆模型} \label{attention_lstm2}
\end{center}
\end{figure}

LSTM模型作为一种经典的循环神经网络结构，在自然语言处理领域是一种常用的序列建模解决方案。对于标准的长短时记忆模型，它从开始到结束按时间顺序处理当前帧的输入，然后将最后时刻的状态向量作为输入序列的最终矢量表示。长短时记忆模型通过其特有的记忆和遗忘机制来建立输入序列之间的长期依赖关系。但是对于文本序列来说，大部分文本信息并不包含想要的类别信息，这些无用信息的积累会导致模型中的记忆单元可信度下降。我们使用了一种基于注意力机制的长短时记忆新模型来进行文本序列建模\cite{ref_proc13}。
标准长短时记忆模型中的记忆机制可以等价表示为图\ref{attention_lstm1}所示的形式。我们把标准的记忆机制分解为两个模块，一个是不随时间变化的本地固有记忆，另一个是随着时间变化不断修改的瞬时记忆。如图\ref{attention_lstm1}所示，本地固有记忆可以表示为多个向量构成的参数矩阵，每个向量代表着本地固有记忆中的某些内容；当前输入和前一时刻的状态向量输入到循环单元中时，当前输入和前一时刻状态向量的串联向量会与本地固有记忆矩阵进行逐行比较，得到一系列的相似度，这些相似度连接成新的瞬时记忆，然后按照控制单元的控制机制写入到瞬时记忆单元中，并产生相应的输出。这种使用当前输入与本地固有记忆的相似度来生成瞬时记忆的方式，会导致我们学习的本地固有记忆与输入内容张成的空间趋同，收敛到输入向量空间的一个子集，无法得到更有鉴别性的记忆表示，并导致过拟合问题。
针对长短时记忆模型中的记忆机制存在的问题，我们构造了新的记忆机制。如图\ref{attention_lstm2}所示，我们首先将单元中的本地固有记忆定义为 个聚类中心构成的码本 ，码本中的每个聚类中心 代表着本地固有记忆中的某些内容。每个当前输入的音频帧 与码本中的所有聚类中心相关联，并计算出距离 最近的码字 。为了使得本地固有记忆与输入内容保持足够的差异性，我们使用 作为当前输入对本地固有记忆的信息增益，并由此产生用于后续处理的瞬时记忆。在这样的记忆机制下，瞬时记忆产生方式与标准长短时记忆模型截然不同。我们首先根据当前输入和前一帧的状态向量预测一下当前时刻的输入内容对应着本地固有记忆中的哪一个码字，然后选取最相似的码字与当前输入内容进行比较，并将其差值作为新的瞬时记忆传输到后面的控制单元中去。

\subsection{中文预训练BERT-wwm}
哈工大讯飞联合实验室发布基于全词覆盖（Whole Word Masking）的中文BERT预训练模型\cite{ref_proc6}，在多个中文数据集上得到了较好的结果，覆盖了句子级到篇章级任务，是BERT的升级版本，主要更改了原预训练阶段的训练样本生成策略。简单来说，原有基于WordPiece的分词方式会把一个完整的词切分成若干个词缀，在生成训练样本时，这些被分开的词缀会随机被[MASK]替换。在全词Mask中，如果一个完整的词的部分WordPiece被[MASK]替换，则同属该词的其他部分也会被[MASK]替换，即全词Mask。

\subsection{单模型及融合实验结果}
我们使用上面的基础层构建了八个单模型，分别为：

\begin{enumerate}
    \item CRF模型
    \item LSTM模型，
    \item LSTM+CRF模型，
    \item CNN+CRF模型，
    \item Self\_Attention模型，
    \item Ucnn模型，
    \item Regressive（WaveNet）模型，
    \item AttentionLSTM模型
\end{enumerate}

另外我们又使用了哈工大讯飞联合实验室的预训练模型WWM以及谷歌的多语种预训练模型Multi，用于初始化，上层使用LSTM+CRF，构建第九个和第十个模型。十个模型的性能结果如表\ref{tab2}所示（训练集场景一900条数据，开发集场景二100条数据）。

我们发现各个单模型预测结果的侧重点会有不同，有很强的多样性，最后我们使用投票的方式进行模型融合，准确率和召回率都有较大提升，如表\ref{tab2}所示。

\begin{table}
\begin{center}
\caption{各个单模型及融合性能}\label{tab2}
\begin{tabular}{|l|l|l|l|}
\hline
模型 &  准确率 & 召回率 & f1值\\
\hline
CRF & 0.7489 & 0.7068 & 0.7272\\
LSTM & 0.7532 & 0.6987 & 0.7250\\
LSTM+CRF & 0.7764 & 0.7670 & 0.7717\\
CNN+CRF & 0.7290 & 0.7349 & 0.7320\\
Self\_Attention & 0.7773 & 0.7429 & 0.7597\\
Ucnn & 0.7338 & 0.7309 & 0.7323\\
Regressive(WaveNet) & 0.7450 & 0.7630 & 0.7539\\
AttentionLSTM & 0.7510 & 0.7510 & 0.7510\\
WWM & 0.7125 & 0.7068 & 0.7096\\
Multi & 0.7551 & 0.7309 & 0.7428\\
融合 & 0.8132 & 0.7871 & \bfseries 0.8000\\
\hline
\end{tabular}
\end{center}
\end{table}

\section{阅读理解模型}

\begin{figure}
\begin{center}
\includegraphics[width=0.5\textwidth]{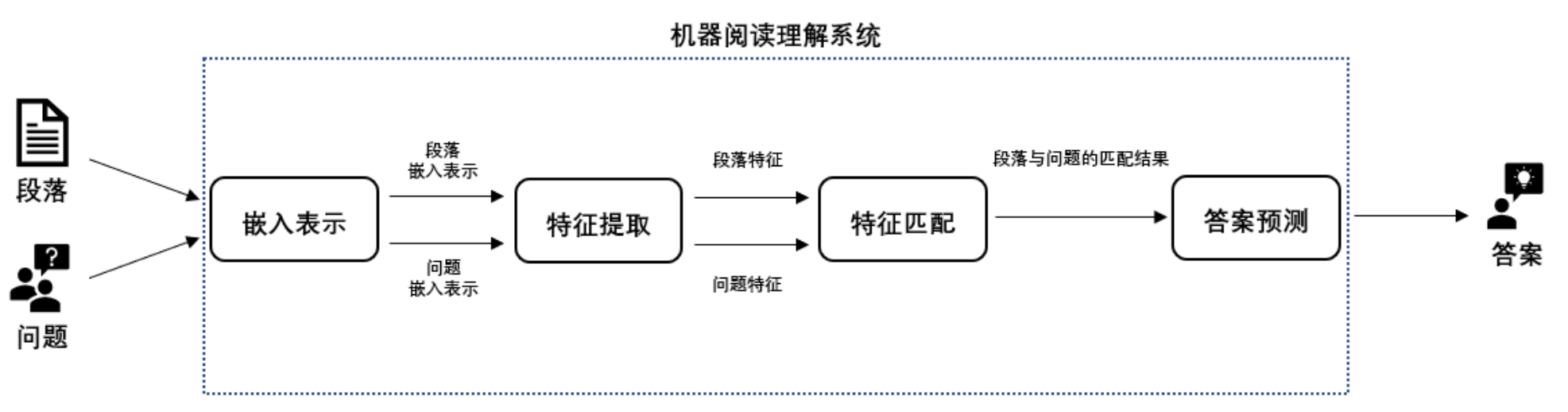}
\caption{机器阅读理解系统的一般流程} \label{fig10}
\end{center}
\end{figure}

此外，我们还尝试了用机器阅读理解的思路来解决病历属性抽取问题。一般来说，机器阅读理解系统的流程如图\ref{fig10}所示：

在本任务中，属性本身可以看作是一个问题，属性值对应的就是从病历文本中找到的输入问题的答案。我们根据这三种属性分别设计了三个问题：原发部位->”原发部位？”、病灶大小->”原发部位的病灶大小是？”、转移部位->”原发部位的转移部位是？”。由于不好事先确定每种属性的个数，我们暂时假设每个问题最多只有一个答案。类似于Boundary Model\cite{ref_proc4}，我们采用的模型的输出是原文中答案的起始位置和答案的终止位置，两个位置之间的文本将会作为最终答案。

不过，若是预测的起始位置和终止位置的概率之和小于一定的阈值，模型也将会舍弃这个结果，选择拒绝回答，也就是说病历文本中不含有该问题对应的属性。我们在场景一900份数据上进行微调，并以场景二100份数据为测试集，在一定的参数设置下，得到的性能如下表\ref{tab3}所示：

\begin{table}
\begin{center}
\caption{阅读理解模型性能}\label{tab3}
\begin{tabular}{|l|l|l|l|}
\hline
模型 &  准确率 & 召回率 & f1值\\
\hline
原发部位 & 0.6438 & 0.6104 & 0.6267\\
病灶大小 & 0.8182 & 0.5806 & 0.6792\\
转移部位 & 0.6909 & 0.2695 & 0.3878\\
所有 & 0.6807 & 0.4137 & 0.5163\\
\hline
\end{tabular}
\end{center}
\end{table}

可以看到，预测的三种属性的的召回率要明显低于其对应的准确率。这是因为病历文本中的属性值往往对应多个，我们的假设造成了模型只会给出一种答案，因此这种机器阅读理解模型的性能还有很大的提升空间。然而，在没有事先确定属性值个数的情况下，如何能够让模型根据属性对应的问题给出多种属性值，仍然是一个非常值得研究的问题，我们希望这种思路能够起到抛砖引玉的效果。

\section{结论}
我们验证序列标注模型可以解决医疗文本属性抽取任务，除了传统的业界比较认可的序列标注模型——lstm+crf的之外，我们尝试了多种序列标注模型（包括CNN,UCNN,self\_attention,WaveNet等），都可以达到一定效果。通过多模型增加系统多样性，融合能提升系统的整体性能。另外我们也尝试使用阅读理解模型解决医疗文本属性抽取任务，虽然不如序列标注模型，但也提供了一种解决问题的思路。
\clearpage
\end{CJK*}

\end{document}